\documentclass[runningheads]{llncs}
\usepackage{graphicx}
\usepackage{amsmath,amssymb} 
\usepackage{bbm}
\usepackage{hyperref}
\usepackage{booktabs}
\usepackage{tabularx}
\usepackage{color}
\usepackage{xspace}
\usepackage{subcaption}
\usepackage[width=122mm,left=12mm,paperwidth=146mm,height=193mm,top=12mm,paperheight=217mm]{geometry}

\usepackage{threeparttable}
\usepackage{array}
\newcolumntype{C}[1]{>{\centering\let\newline\\\arraybackslash\hspace{0pt}}m{#1}}
\renewcommand{\arraystretch}{1.2}
\usepackage{caption} 
 \makeatletter
 \DeclareRobustCommand\onedot{\futurelet\@let@token\@onedot}
 \def\@onedot{\ifx\@let@token.\else.\null\fi\xspace}
  
 \def\ie{i.e\onedot}

 \makeatother

\graphicspath{{./fig/}{./fig/plots/}}

\newcommand{\pnorm}[2]{\ensuremath{|| #1 ||_{#2}}}

\newcommand{\twoNorm}[1]{\pnorm{#1}{2}}

\newcommand{\bs}[1]{\ensuremath{\boldsymbol{#1}}}

\DeclareMathOperator* {\argmax}{arg\, max}

\begin{document}

\pagestyle{headings}
\mainmatter
\definecolor{orange}{rgb}{1,0.5,0}
\definecolor{darkgreen}{rgb}{0,0.5,0}
\definecolor{red}{rgb}{1,0,0}

\newcommand{\correct}[1]{\textcolor{red}{#1}}

\title{Learning Visual Question Answering by Bootstrapping Hard Attention}

\titlerunning{Learning Visual Question Answering by Bootstrapping Hard Attention}

\author{Mateusz Malinowski\and
Carl Doersch \and
Adam Santoro \and
Peter Battaglia
}
%
\authorrunning{M. Malinowski, C. Doersch, A. Santoro and P. Battaglia}
%

\institute{DeepMind, London, United Kingdom \\
}

\maketitle

\begin{abstract}
Attention mechanisms in biological perception are thought to select subsets of perceptual information for more sophisticated processing which would be prohibitive to perform on all sensory inputs. In computer vision, however, there has been relatively little exploration of \emph{hard} attention, where some information is selectively ignored, in spite of the success of \emph{soft} attention, where information is re-weighted and aggregated, but never filtered out.
Here, we introduce 
a new approach for hard attention and find it achieves
very competitive performance on a recently-released visual question answering datasets, equalling and in some cases surpassing similar soft attention architectures while entirely ignoring some features.
Even though the hard attention mechanism is thought to be non-differentiable, 
we found that the feature magnitudes correlate with semantic relevance, and provide a useful signal for our mechanism's attentional selection criterion.
Because hard attention selects important features of the input information, it can also be more efficient than analogous soft attention mechanisms. This is especially important for recent approaches that use \emph{non-local pairwise} operations, whereby computational and memory costs are quadratic in the size of the set of features.

\end{abstract}

\section{Introduction}

\begin{figure*}[tp]
\begin{center}

\includegraphics[width=1\linewidth]{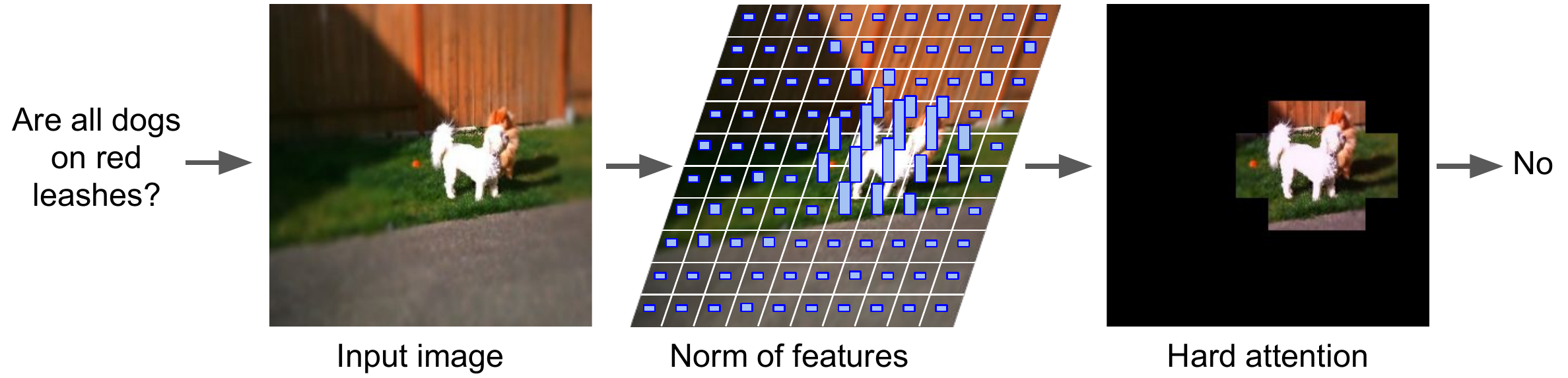}

\end{center}
\caption{Given a natural image and a textual question as input, our Visual QA architecture outputs an answer. It uses a hard attention mechanism that selects only the important visual features for the task for further processing. We base our architecture on the premise that the norm of the visual features correlates with their relevance, and that those feature vectors with high magnitudes correspond to image regions which contain important semantic content.}

\label{fig:teaser}
\end{figure*}

\label{sec:intro}
Visual attention is instrumental to many aspects of complex visual reasoning in humans \cite{ccukur2013attention,sheinberg2001noticing}. 
For example, when asked to identify a dog's owner among a group of people, 
the human visual system adaptively allocates greater computational resources to processing visual information associated with the dog and potential owners, versus other aspects of the scene.
The perceptual effects can be so dramatic that prominent entities may not even rise to the level of awareness when the viewer is attending to other things in the scene \cite{simons2005change,mack1998inattentional,simons1999gorillas}. 
Yet attention has not been a transformative force in computer vision, possibly because many standard computer vision tasks like detection, segmentation, and classification do not involve the sort of complex reasoning which attention is thought to facilitate.

Answering detailed questions about an image is a type of task which requires more sophisticated patterns of reasoning, and there has been a rapid recent proliferation of computer vision approaches for tackling the \textit{visual question answering} (Visual QA) task~\cite{malinowski14nips,agrawal2017don}. 
Successful Visual QA architectures must be able to handle many objects and their complex relations while also integrating rich background knowledge, and attention has emerged as a promising strategy for achieving good performance~\cite{agrawal2017don,xu2015ask,yang2015stacked,fukui16emnlp,perez2017film,kazemi2017show,teney2017tips,ilievski2016fda}.

We recognize a broad distinction between types of attention in computer vision and machine learning -- {\em soft} versus {\em hard} attention.
Existing attention models \cite{agrawal2017don,xu2015ask,yang2015stacked,fukui16emnlp}
are predominantly based on soft attention, in which all information is adaptively re-weighted before being aggregated.
This can improve accuracy by isolating important information and avoiding interference from unimportant information. Learning becomes more data efficient as the complexity of the interactions among different pieces of information reduces; this, loosely speaking, allows for more unambiguous credit assignment.

By contrast, hard attention, in which only a subset of information is selected for further processing, is much less widely used. Like soft attention, it has the potential to improve accuracy and learning efficiency by focusing computation on the important parts of an image. 
But beyond this, it offers better computational efficiency because it only fully processes the information deemed most relevant.
However, there is a key downside of hard attention within a gradient-based learning framework, such as deep learning: because the choice of which information to process is discrete and thus non-differentiable, gradients cannot be backpropagated into the selection mechanism to support gradient-based optimization. 
There have been various efforts to address this shortcoming in visual attention \cite{mnih2014recurrent}, attention to text \cite{gulcehre2016dynamic}, and more general machine learning domains \cite{bengio2013estimating,jang2016categorical,maddison2016concrete}, but this is still a very active area of research.

Here we explore a simple approach to hard attention that bootstraps on an interesting phenomenon \cite{olah2018building} in the feature representations of convolutional neural networks (CNNs): learned features often carry an easily accessible signal for hard attentional selection. In particular, selecting those feature vectors with the greatest $L2$-norm values proves to be a heuristic that can facilitate hard attention -- and provide the performance and efficiency benefits associated with -- without requiring specialized learning procedures (see \autoref{fig:teaser}). This attentional signal results indirectly from a standard supervised task loss, and does not require explicit supervision to incentivize norms to be proportional to object presence, salience, or other potentially meaningful measures \cite{olah2018building,oliva2003top}. 

We rely on a canonical Visual QA pipeline \cite{agrawal2017don,yang2015stacked,ren2015image,antol2015vqa,malinowski2017ask,santoro2017simple} augmented with a hard attention mechanism that uses the $L2$-norms of the feature vectors to select subsets of the information for further processing. The first version, called the Hard Attention Network (HAN), selects a fixed number of feature vectors by choosing those with the top norms. The second version, called the Adaptive Hard Attention Network (AdaHAN), selects a variable number of feature vectors that depends on the input.
Our results show that our algorithm can actually outperform comparable soft attention architectures on a challenging Visual QA task.
This approach also produces interpretable hard attention masks, where the image regions which correspond to the selected features often contain semantically meaningful information, such as coherent objects.
We also show strong performance when combined with a form of 
non-local pairwise model \cite{wang2017non,santoro2017simple,vaswani2017attention,shaw2018self}.  This algorithm computes features over pairs of input features and thus scale quadratically with number of vectors in the feature map, highlighting the importance of feature selection.

\section{Related Work}
\label{sec:related_work}
Visual question answering, 
or more broadly the Visual Turing Test, asks ``Can machines understand a visual scene only from answering questions?'' \cite{malinowski14nips,antol2015vqa,malinowski14visualturing,malinowski2015hard,malinowski2017towards,geman2015visual}. 
Creating a good Visual QA dataset has proved non-trivial: biases in the early datasets~\cite{malinowski14nips,ren2015image,antol2015vqa,gao2015you} rewarded algorithms for exploiting spurious correlations, rather than tackling the reasoning problem head-on \cite{agrawal2017don,johnson2017clevr,goyal2017making}.
Thus, we focus on  the recently-introduced VQA-CP \cite{agrawal2017don} and CLEVR \cite{johnson2017clevr} datasets, which aim to reduce the dataset biases, providing a more difficult challenge for rich visual reasoning.

One of the core challenges of Visual QA is the problem of {\em grounding language}: that is, associating the meaning of a language term with a specific perceptual input \cite{harnad1990symbol}.  
Many works have tackled this problem \cite{guadarrama2013grounding,kong2014you,karpathy15cvpr,rohrbach2016grounding}, enforcing that language terms be grounded in the image. 
In contrast, our algorithm does not directly use correspondence between modalities to enforce such grounding but instead relies on learning to find a discrete representation that captures the required information from the raw visual input, and question-answer pairs.

The most successful Visual QA architectures build multimodal representations with a combined CNN+LSTM architecture~\cite{ren2015image,gao2015you,malinowski2015ask}, and recently have begun including attention mechanisms inspired by soft and hard attention for image captioning~\cite{xu15icml}. 
However, only soft attention is used in the majority of Visual QA works~\cite{agrawal2017don,xu2015ask,yang2015stacked,fukui16emnlp,perez2017film,kazemi2017show,xiong16dynamic,de2017modulating,zhu16cvpr,chen2015abc,shih2015look,gulcehre2018hyperbolic,hudson2018compositional,mascharka2018transparency,kim2017structured,explainable2018eccv}. 
In these architectures, a full-frame CNN representation is used to compute a spatial weighting (attention) over the CNN grid cells. 
The visual representation is then the weighted-sum of the input tensor across space. 

The alternative is to select CNN grid cells in a discrete way, but due to many challenges in training non-differentiable architectures, such hard attention alternatives are severely under-explored. 
Notable exceptions include \cite{malinowski14nips,teney2017tips,ilievski2016fda,mokarian2016mean,tommasi2016solving,desta2018object}, but these run state-of-the-art object detectors or proposals to compute the hard attention maps. 
We argue that relying on such external tools is fundamentally limited: it requires costly annotations, and cannot easily adapt to new visual concepts that aren't previously labeled.
Outside Visual QA and captioning, some prior work in vision has explored limited forms of hard attention.  One line of work on discriminative patches builds a representation by selecting some patches and ignoring others, which has proved useful for object detection and classification~\cite{singh2012unsupervised,doersch2013mid,juneja2013blocks}, and especially visualization~\cite{doersch2012makes}.  
However, such methods have recently been largely supplanted by end-to-end feature learning for practical vision problems.  In deep learning, spatial transformers \cite{jaderberg2015spatial} are one method for selecting an image regions while ignoring the rest, although these have proved challenging to train in practice.
Recent work on compressing neural networks (e.g.~\cite{mallya2017packnet}) uses magnitudes to remove weights of neural networks. However it  prunes  permanently  based  on weight magnitudes, not dynamically based on activation norms, and has no direct connection to hard-attention or Visual QA.

Attention has also been studied outside of vision. While the focus on soft attention predominates these works as well, there are a few examples of hard attention mechanisms and other forms of discrete gating~\cite{mnih2014recurrent,gulcehre2016dynamic,bengio2013estimating,jang2016categorical,maddison2016concrete}. In such works the decision of \emph{where to look} is seen as a discrete variable that had been optimized either by reinforce loss or various other approximations (e.g. straight-through). However, due to the high variance of these gradients, learning can be inefficient, and soft attention mechanisms usually perform better.  

\section{Method}
\label{sec:method}

\begin{figure*}[tp]
\begin{center}

\includegraphics[width=\linewidth]{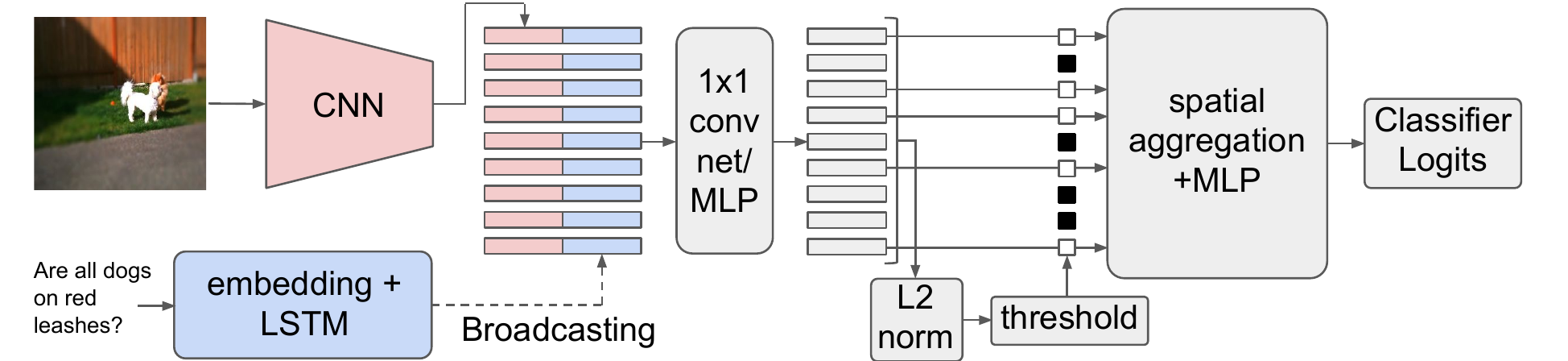}

\end{center}
\caption{Our hard attention replaces commonly used soft attention mechanism. Otherwise, we follow the canonical Visual QA pipeline \cite{agrawal2017don,yang2015stacked,ren2015image,antol2015vqa,malinowski2017ask,santoro2017simple}. Questions and images are  encoded into their vector representations. Next, the spatial encoding of the visual features is unraveled, and the question embedding is broadcasted and concatenated (or added) accordingly to form a multimodal representation of the inputs. Our attention mechanism selectively chooses a subset of the multimodal vectors that are next aggregated and processed by the answering module.}

\label{fig:vqa_encoder_decoder_view}
\end{figure*}
Answering questions about images is often formulated in terms of predictive models \cite{malinowski2017ask}. These architectures  maximize a conditional distribution over answers $a$, given questions $q$ and images $x$: 
\begin{equation}
\label{eq:vqa_formulation}
\hat{a}=\argmax_{a \in \mathcal{A}}p(a|x,q)
\end{equation}
where $\mathcal{A}$ is a countable set of all possible answers. 
As is common in question answering \cite{agrawal2017don,yang2015stacked,ren2015image,antol2015vqa,malinowski2017ask}, the question is a sequence of words $q =\left[q_1,...,q_n\right]$, while the output is reduced to a classification problem between a set of common answers (this is limited compared to approaches that generate answers \cite{malinowski2015ask}, but works better in practice). 
Our architecture for learning a mapping from image and question, to answer, is shown in~\autoref{fig:vqa_encoder_decoder_view}.
We encode the image with a CNN~\cite{krizhevsky2012imagenet} (in our case, a pre-trained ResNet-101~\cite{he2015deep}, or a small CNN trained from scratch),
and encode the question to a fixed-length vector representation with an LSTM \cite{hochreiter97nc}. 
We compute a combined representation by copying the question representation to every spatial location in the CNN, and concatenating it with (or simply adding it to) the visual features, like previous work~\cite{agrawal2017don,yang2015stacked,ren2015image,antol2015vqa,malinowski2017ask,santoro2017simple}. 
After a few layers of combined processing, we apply attention over spatial locations, 
following previous works which often apply soft attention mechanisms \cite{agrawal2017don,xu2015ask,yang2015stacked,fukui16emnlp} at this point in the architecture.
Finally, we aggregate features, using either sum-pooling, or relational \cite{santoro2017simple,vaswani2017attention,battaglia2018relational} modules.
We train the whole network end-to-end with a standard logistic regression loss over answer categories.

\subsection{Attention Mechanisms}
Here, we describe prior work on soft attention, and our approach to hard attention.
\newline
\textbf{Soft Attention.}
In most prior work, soft attention is implemented as a weighted mask over the spatial cells of the CNN representation.
Let $\bs{x} := CNN(x), \bs{q} := LSTM(q)$ for image $x$ and question $q$.  
We compute a weight $w_{ij}$ for every $\bs{x}_{ij}$ (where $i$ and $j$ index spatial locations), using a neural network that takes $\bs{x}_{ij}$ and $\bs{q}$ as input.  
Intuitively, weight $w_{ij}$ measures the ``relevance'' of the cell to the input question.
$w$ is nonnegative and normalized to sum to 1 across the image (generally with softmax).
Thus, $w$ is applied to the visual input via 
$\hat{\bs{h}_{ij}} := w_{ij} \bs{x}_{ij}$ 
to build the multi-modal representation.
This approach has some advantages, including conceptual simplicity and  differentiability.
The disadvantage is that the weights, in practice, are never 0.
Irrelevant background can affect the output, no features can be dropped from potential further processing, and credit assignment is 
still
challenging.
\newline
\textbf{Hard Attention.}
Our main contribution is a new mechanism for hard attention. It produces a binary mask over spatial locations, which determines which features are passed on to further processing.
We call our method the Hard Attention Network (HAN).
The key idea is to use the $L2$-norm of the activations at each spatial location as a proxy for relevance at that location. 
The correlation between $L2$-norm and relevance is an emergent property of the trained CNN features, which requires no additional constraints or objectives.
\cite{olah2018building} recently found something related: in an ImageNet-pretrained representation of an image of a cat and a dog, the largest feature norms appear above the cat and dog face, even though the representation was trained purely for classification.
Our architecture bootstraps on this phenomenon without explicitly training the network to have it.

As above, let $\bs{x}_{ij}$ and $\bs{q}$ be a CNN cell at the spatial position $i,j$, and a question representation respectively. 
We first embed $\bs{q} \in \mathbb{R}^q$ and $\bs{x} \in \mathbb{R}^x$ into two feature spaces that share the same dimensionality $d$, \ie,
\begin{align}
\bs{\hat{x}} &:= CNN^{1 \times 1}(\bs{x}; \theta_x) \label{eq:emb_x} \in \mathbb{R}^{w \times h \times d} \\
\bs{\hat{q}} &:= MLP(\bs{q}; \theta_q) \label{eq:emb_q}  \in \mathbb{R}^d
\end{align}
where $CNN^{1 \times 1}$ stands for a $1 \times 1$ convolutional network and $MLP$ stands for a multilayer perceptron. 
We then combine both the convolutional image features with the question features into a shared \emph{multimodal} embedding by first broadcasting the question features to match the $w \times d$ shape of the image feature map, and then performing element-wise addition (1x1 conv net/MLP in Figure~\ref{fig:vqa_encoder_decoder_view}):
\begin{align}
\bs{m}_{ij} := \bs{\hat{x}}_{ij} \oplus \bs{\hat{q}}\text{\ , where }\bs{m} := \left[\bs{m}_{ij}\right]_{ij} \in \mathbb{R}^{w \times h \times d}
\end{align}
Element-wise addition keeps the dimensionality of each input, as opposed to concatenation, yet is still effective \cite{kazemi2017show,malinowski2017ask}. 
Next, we compute the \emph{presence vector},
$\bs{p} := \left[p_{ij}\right]_{ij} \in \mathbb{R}^{w \times h}$ which measures the relevance of entities given the question:
\begin{equation}
\label{eq:attention_measure}
 p_{ij} := \twoNorm{\bs{m}_{ij}} \in \mathbb{R}
\end{equation}
where $\twoNorm{\cdot}$ denotes $L2$-norm.
To select $k$ entities from $\bs{m}$ for further processing, the indices of the top $k$ entries in $\bs{p}$, denoted $\bs{l}=\left[l_1, \dots, l_k\right]$ are used to form $\hat{\bs{m}}^k = \left[\bs{m}_{l_1}, ..., \bs{m}_{l_k}\right] \in \mathbb{R}^{k \times d}$.

This set of features is passed to the decoder module and gradients will flow back to the weights of the CNN/MLP through the selected features.   
Our assumption is that important outputs of the CNN/MLP will tend to grow in norm, and therefore are likely to be selected. Intuitively if less useful features are selected, the gradients will push the norm of these features down, making them less likely to be selected again.
But there is nothing in our framework which explicitly incorporates this behavior into a loss.
Despite its simplicity, our experiments (Section \ref{sec:results}) show the HAN is very competitive with canonical soft attention \cite{yang2015stacked} while also offering interpretability and efficiency.

Thus far, we have assumed that we can fix the number of features $k$ that are passed through the attention mechanism.  
However, it is likely that different questions require different spatial support within the image. 
Thus, we also introduce a second approach which adaptively chooses the number of entities to attend to (termed Adaptive-HAN, or AdaHAN) as a function of the inputs, rather than using a fixed $k$. 
The key idea is to make the presence vector $\bs{p}$ (the norm of the embedding at each spatial location) ``compete'' against a threshold $\tau$.
However, since the norm is unbounded from above, to avoid trivial solutions in which the network sets the presence vector very high and selects all entities, we apply a softmax operator to $\bs{p}$. We put both parts into the competition by only selecting those elements of $\bs{m}$ whose presence values exceed the threshold,
\begin{align}
\label{eq:competition}
\bs{\hat{m}}^k = \left[\bs{m}_{l_1}, ..., \bs{m}_{l_k}\right] \in \mathbb{R}^{k \times d}
\text{\ , where } \{l_i : \text{softmax}(\bs{p}_{l_i}) > \tau \}
\end{align}
Note that due to the properties of softmax, the competition is encouraged not only between both sides of the inequality, but also between the spatially distributed elements of the presence vector $\bs{p}$.
Although $\tau$ could be chosen through the hyper-parameter selection, we decide to use $\tau := \frac{1}{w\cdot h}$ where $w$ and $h$ are spatial dimensions of the  input vector $x_{ij}$. 
Such value for $\tau$ has an interesting interpretation. If each spatial location of the input were equally important, we would sample the locations from a uniform probability distribution $p(\cdot) := \tau = \frac{1}{w\cdot h}$. This is equivalent to a probability distribution induced by the presence vector of a neural network with uniformly distributed spatial representation, \ie $\tau = \text{softmax}(\bs{p}_{\text{uniform}})$,  and hence the trained network with the presence vector $\bs{p}$ has to ``win'' against the $\bs{p}_{\text{uniform}}$ of the random network in order to select right input features by shifting the probability mass accordingly.
It also naturally encourages higher selectivity as the increase in the probability mass at one location would result in decrease in another location.

In contrast to the commonly used soft-attention mechanism, our approaches do not require extra learnable parameters. HAN requires a single extra but interpretable hyper-parameter: a fraction of input cells to use, which trades off speed for accuracy.  AdaHAN requires no extra hyper-parameters.

\subsection{Feature Aggregation}
\textbf{Sum Pooling.} A simple way to reduce the set of feature vectors after attention is to sum pool them into a constant length vector.
In the case of a soft attention module with an attention weight vector $w$, it is straightforward to compute a pooled vector as $\sum_{ij} w_{ij} \bs{x}_{ij}$.  
Given features selected with hard attention, an analogous pooling can be written as  $\sum_{\kappa=1}^{k}\bs{m}_{l_\kappa}$.
\newline
\textbf{Non-local Pairwise Operator.}
To improve on sum pooling, we explore an approach which performs reasoning through non-local and pairwise computations, one of a family of similar architectures which has shown promising results for question-answering and video understanding \cite{santoro2017simple,wang2017non,vaswani2017attention}.
An important aspect of these non-local pairwise methods is that the computation is quadratic in the number of features, and thus hard attention can provide significant computational savings.
Given some set of embedding vectors (such as the spatial cells of the output of a convolutional layer) $\bs{x}_{ij}$, one can use three simple linear projections to produce a matrix of queries, $\bs{q}_{ij} := \bs{W}_q \bs{x}_{ij}$, keys, $\bs{k}_{ij} := \bs{W}_k \bs{x}_{ij}$, and values, $\bs{v}_{ij}=\bs{W}_v\bs{x}_{ij}$ at each spatial location. 
Then, for each spatial location $i,j$, we compare the query $q_{ij}$ with the keys at all other locations, and sum the values $v$ weighted by the similarity. 
Mathematically, we compute 
\begin{align}
\tilde{\bs{x}}_{lk} = \sum_{ij} \text{softmax}\left(\bs{q}_{lk}^T\bs{k}_{ij}\right)\bs{v}_{ij}
\end{align}
Here, the softmax operates over all $i,j$ locations. The final representation of the input is computed by summarizing all $\tilde{\bs{x}}_{lk}$ representations, e.g. we use sum-pooling to achieve this goal.
Thus, the mechanism computes non-local \cite{wang2017non} pairwise relations between embeddings, independent of spatial or temporal proximity.  
The separation between keys, queries, and values allows semantic information about each object to remain separated from the information that binds objects together across space.
The result is an effective, if somewhat expensive, spatial reasoning mechanism.
Although expensive, similar mechanism has been shown useful in various tasks, from synthetic visual question \cite{santoro2017simple}, to machine translation \cite{vaswani2017attention}, to video recognition \cite{wang2017non}.
Hard attention can help to reduce the set of comparisons that must be considered, and thus we aim to test whether the features selected by hard attention are compatible with this operator.

\section{Results}
\label{sec:results}
To show the importance of hard attention for Visual QA, we first compare HAN to existing soft attention (SAN) architectures on VQA-CP v2, and exploring the effect of varying degrees of hard attention by directly controlling the number of attended spatial cells in the convolutional map. 
We then examine AdaHAN, which adaptively chooses the number of attended cells, and briefly investigate the effect of network depth and pretraining.
Finally, we present qualitative results, and also provide results on CLEVR to show the method's generality.

\subsection{Datasets}
\textbf{VQA-CP v2.} This dataset~\cite{agrawal2017don}
consists of about 121K (98K) images, 438K (220K) questions, and 4.4M (2.2M) answers in the train (test) set; and it is created so that the distribution of the answers between train and test splits differ, and hence the models cannot excessively rely on the language prior \cite{agrawal2017don}. As expected, \cite{agrawal2017don} show that performance of all Visual QA approaches they tested drops significantly between train to test sets. 
The dataset provides a standard train-test split, and also breaks questions into different question types: those where the answer is yes/no, those where the answer is a number, and those where the answer is something else. Thus, we report accuracy on each question type as well as the overall accuracy for each network architecture.
\newline
\textbf{CLEVR.} This synthetic dataset~\cite{johnson2017clevr} consists of 100K images of 3D rendered objects like spheres and cylinders, and roughly 1m questions that were automatically generated with a procedural engine.  While the visual task is relatively simple, solving this dataset requires reasoning over complex relationships between many objects.

\subsection{Effect of Hard Attention}
We begin with the most basic hard attention architecture, which applies hard attention and then does sum pooling over the attended cells, followed by a small MLP.
For each experiment, we take the top $k$ cells, out of $100$, according to our $L2$-norm criterion, where $k$ ranges from $16$ to $100$ (with $100$, there is no attention, and the whole image is summed).  
Results are shown in the top of Table~\ref{table:han-mhsa-attention_numbers}.  
Considering that the hard attention 
selects only a subset of the input cells, 
we might expect that the algorithm would lose important information and be unable to recover.
In fact, however, the performance is almost the same with less than half of the units attended.
Even with just $16$ units, the performance loss is less than 1\%, suggesting that hard attention is quite capable of capturing the important parts of the image.

\begin{table}
\begin{center}
\begin{tabular}{llrrrr}
\toprule
 & Percentage\,\, & Overall\,\, & Yes/No\,\, & Number\,\, & Other \\
 & of cells\,\,  & & & & \\
 \cmidrule(ll){1-2}\cmidrule(lr){3-6}
 HAN+sum & 16\% & 26.99 & 40.53 & 11.38 & 24.15 \\
 HAN+sum & 32\% & 27.43 & 41.05 & 11.38 & 24.68 \\
 HAN+sum & 48\% & 27.94 & 41.35 & 11.93 & 25.27 \\
 HAN+sum & 64\% & 27.80 & 40.74 & 11.29 & 25.52 \\
 sum & 100\% & 27.96 & 43.23 & 12.09 & 24.29 \\ 
\midrule
 HAN+pairwise & 16\% & 26.81 & 41.24 & 10.87 & 23.61 \\
 HAN+pairwise & 32\% & 27.45 & 40.91 & 11.48 & 24.75 \\
 HAN+pairwise & 48\% & 28.23 & 41.23 & 11.40 & 25.98 \\
 Pairwise & 100\%  & 28.06 & 44.10 & 13.20 & 23.71 \\
 \midrule
  SAN \cite{agrawal2017don,yang2015stacked} & - & 24.96 & 38.35 & 11.14 & 21.74 \\
  SAN (ours) & - & 26.60 & 39.69 & 11.25 & 23.92 \\
  SAN+pos (ours) & - & 27.77 & 40.73 & 11.31 & 25.47 \\
  \midrule
  GVQA \cite{agrawal2017don} & - & 31.30 & 57.99 & 13.68 & 22.14 \\
\bottomrule
\end{tabular}
\end{center}
\caption{
Comparison between different number of attended cells (percentage of the whole input), and aggregation operation. We consider a simple summation, and non-local pairwise computations as the aggregation tool. 
}
\label{table:han-mhsa-attention_numbers}
\end{table}

The fact that hard attention can work is interesting itself, but it should be especially useful for models that devote significant processing to each attended cell.  
We therefore repeat the above experiment with the non-local pairwise aggregation mechanism described in \autoref{sec:method}, which computes activations for every pair of attended cells, and therefore scales quadratically with the number of attended cells.
These results are shown in the middle of Table~\ref{table:han-mhsa-attention_numbers}, where we can see that hard attention (48 entitties) actually boosts performance over an analogous model without hard attention.

Finally, we compare standard soft attention baselines in the bottom of \autoref{table:han-mhsa-attention_numbers}.
In particular, we include previous results using a basic soft attention network~\cite{agrawal2017don,yang2015stacked}, as well as our own re-implementation of the soft attention pooling algorithm presented in \cite{agrawal2017don,yang2015stacked} with the same features used in other experiments.  
Surprisingly, soft attention does not outperform basic sum pooling, even with careful implementation that outperforms the previously reported results with the same method on this dataset; in fact, it performs slightly worse. 
The non-local pairwise aggregation performs better than SAN on its own, although the best result includes hard attention.
Our results overall are somewhat worse than the state-of-the-art~\cite{agrawal2017don}, but this is likely due to several architectural decisions not included here, such as a split pathway for different kinds of questions, special question embeddings, and the use of the question extractor.

\begin{table*}
\begin{center}
\begin{tabular}{llrrrr}
\toprule
 & Percentage\,\, & Overall\,\, & Yes/No\,\, & Number\,\, & Other \\
 & of cells & & & \\
 \cmidrule(l){1-1}\cmidrule(lr){2-5}
 AdaHAN+sum & 25.66\% & 27.40 & 40.70 & 11.13 & 24.86 \\
 AdaHAN+pairwise  & 32.63\% & 28.65 & 52.25 & 13.79 & 20.33 \\
 \midrule
 HAN+sum  & 32\% & 27.43 & 41.05 & 11.38 & 24.68 \\
 HAN+sum  & 48\% & 27.94 & 41.35 & 11.93 & 25.27 \\
 HAN+pairwise  & 32\% & 27.45 & 40.91 & 11.48 & 24.75 \\
 HAN+pairwise  & 48\% & 28.23 & 41.23 & 11.40 & 25.98 \\
\bottomrule
\end{tabular}
\end{center}
\caption{
Comparison between different adaptive hard-attention techniques with average number of attended parts,  and aggregation operation. We consider a simple summation, and the non-local pairwise aggregation. 
Since AdaHAN adaptively selects relevant features, based on the fixed threshold $\frac{1}{w*h}$, we report here the average number of attended parts.
}
\label{table:adaptive_han-attention_numbers}
\end{table*}

\subsection{Adaptive hard attention}
Thus far, our experiments have dealt with networks that have a fixed threshold for all images.  
However, some images and questions may require reasoning about more entities than others.
Therefore, we explore a simple adaptive method, where the network chooses how many cells to attend to for each image.
Table~\ref{table:adaptive_han-attention_numbers} shows results, where AdaHAN refers to our adaptive mechanism.  We can see that on average, the adaptive mechanism uses surprisingly few cells: 25.66 out of 100 when sum pooling is used, and 32.63 whenever the non-local pairwise aggregation mechanism is used.
For sum pooling, this is on-par with a non-adaptive network that uses more cells on average (HAN+sum 32); for the non-local pairwise aggregation mechanism, just 32.63 cells are enough to outperform our best non-adaptive model, which uses roughly $50\%$ more cells.
This shows that even very simple methods of adapting hard attention to the image and the question can lead to both computation and performance gains, suggesting that more sophisticated methods will be an important direction for future work.

\subsection{Effects of network depth}
In this section, we briefly analyze an important architectural choice: the number of layers used on top of the pretrained embeddings.  
That is, before the question and image representations are combined, we perform a small amount of processing to ``align'' the information, so that the embedding can easily tell the relevance of the visual information to the question.
Table~\ref{table:han-sum-attention_numbers} shows the results of removing the two layers which perform this function.  
We consistently see a drop of about 1\% without the layers, suggesting that deciding which cells to attend to requires different information than the classification-tuned ResNet is designed to provide.

\begin{table*}[tb]
\begin{center}
\begin{tabular}{lllrrrr}
\toprule
 & Percentage\,\, & Number\,\, & Overall\,\, & Yes/No\,\, & Number\,\, & Other \\
 & of cells\,\, & of layers\,\,  & & & & \\
 \cmidrule(ll){1-3}\cmidrule(lr){4-7}
 HAN+sum & 25\% & 0 & 26.38 & 43.21 & 13.12 & 21.17 \\
 HAN+sum & 50\% & 0 & 26.75 & 41.42 & 10.94 & 23.38 \\
 HAN+sum & 75\%   & 0 & 26.82 & 41.30 & 11.48 & 23.42 \\
\midrule
 HAN+sum & 25\% & 2 & 26.99 & 40.53 & 11.38 & 24.15 \\
 HAN+sum & 50\% & 2 & 27.43 & 41.05 & 11.38 & 24.68 \\
 HAN+sum & 75\% & 2 & 27.94 & 41.35 & 11.93 & 25.27 \\
\bottomrule
\end{tabular}
\end{center}
\caption{
Comparison between different number of the attended 
cells as the percentage of the whole input. The results are reported on VQA-CP v2. The second column denotes the 
percentage of the attended input.
The third column denotes number of layers of the MLP (Equations \ref{eq:emb_x} and \ref{eq:emb_q}).
}
\label{table:han-sum-attention_numbers}
\end{table*}

\subsection{Implementation Details.}
All our models use the same LSTM size $512$ for questions embeddings, and the last convolutional layer of the ImageNet pre-trained ResNet-101 \cite{he2015deep} (yielding 10-by-10 spatial representation, each with $2048$ dimensional cells) for image embedding. We also use MLP with $3$ layers of sizes: $1024, 2048, 1000$, as a classification module. We use ADAM for optimization \cite{kingma2014adam}. We use a distributed setting with two workers computing a gradient over a batch of $128$ elements each. We normalize images by dividing them by their norm. We do not perform a hyper-parameter search as there is no separated validation set available. Instead, we rather choose default hyper-parameters based on our prior experience on Visual QA datasets. We trained our models until we notice a saturation on the training set. Then we evaluate these models on the test set. Our tables show the performance of all the methods wrt. the second digits precision obtained by rounding.

\autoref{table:han-mhsa-attention_numbers} shows SAN's \cite{yang2015stacked} results reported by \cite{agrawal2017don} together with our in-house implementation (denoted as ``ours''). Our implementation has 2 attention hops, $1024$ dimensional multimodal embedding size, a fixed learning rate $0.0001$, and ResNet-101. In these experiments we pool the attended representations by weighted average with the attention weights.
Our in-house implementation of the non-local pairwise mechanism strongly resembles implementations of \cite{wang2017non}, and \cite{vaswani2017attention}. We use $2$ heads, with embedding size $512$.
In \autoref{eq:emb_x} and \autoref{eq:emb_q}, we use $d := 2048$ (the same as dimensionality as the image encoding) and two linear layers with RELU that follows up each layer.

\subsection{Qualitative Results.}
\label{sec:qualitative}
\begin{figure*}[p]
\begin{center}
\begin{tabular}{l@{\ }l@{\ }c@{\ }c@{\ }c@{\ }c}
 \toprule
 & \multicolumn{1}{c}{Are those drinking} & 
\multicolumn{1}{c}{Can cars cross} & 
\multicolumn{1}{c}{What color is} &
\multicolumn{1}{c}{What is building} \\
& \multicolumn{1}{c}{glasses next to} & 
\multicolumn{1}{c}{this bridge?} & 
\multicolumn{1}{c}{her skirt?} & 
\multicolumn{1}{c}{facade made} \\
& \multicolumn{1}{c}{flower pot?} & 
\multicolumn{1}{c}{} & 
\multicolumn{1}{c}{} & 
\multicolumn{1}{c}{from?}
\\\midrule
\rotatebox{90}{AdaHAN+pair.} & \multicolumn{1}{c}{\includegraphics[width=0.2\linewidth]{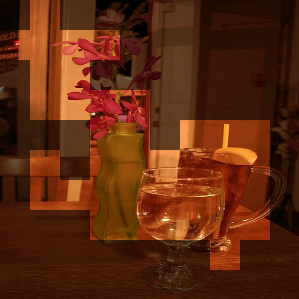}} &
\includegraphics[width=0.2\linewidth]{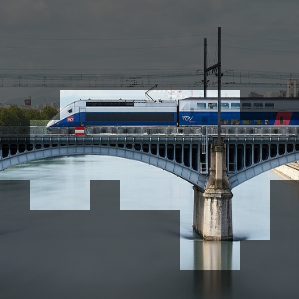} &
\includegraphics[width=0.2\linewidth]{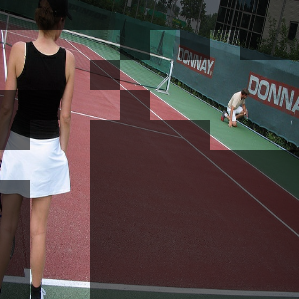} &
\includegraphics[width=0.2\linewidth]{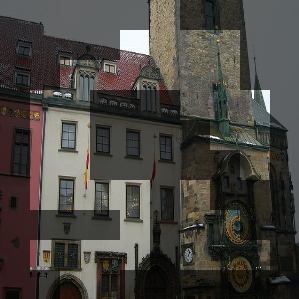}
\\
 & \multicolumn{1}{c}{\textcolor{darkgreen}{yes}} & 
\multicolumn{1}{c}{\textcolor{darkgreen}{no}} & 
\multicolumn{1}{c}{\textcolor{darkgreen}{white}} &
\multicolumn{1}{c}{\textcolor{darkgreen}{brick}}
\\\midrule
\rotatebox{90}{AdaHAN+sum} & \multicolumn{1}{c}{\includegraphics[width=0.2\linewidth]{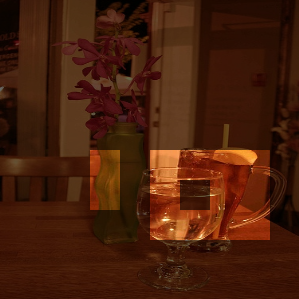}} &
\includegraphics[width=0.2\linewidth]{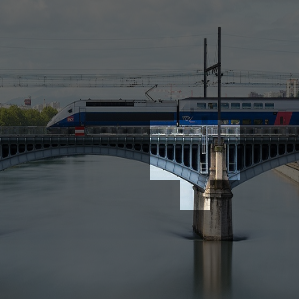} &
\includegraphics[width=0.2\linewidth]{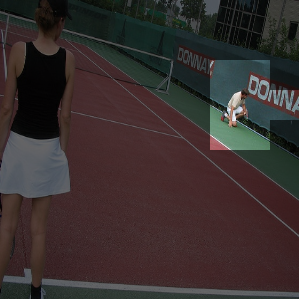} &
\includegraphics[width=0.2\linewidth]{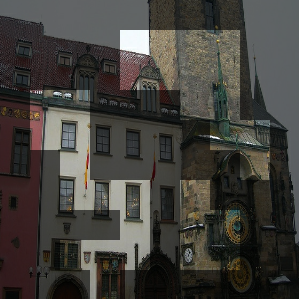}
\\
 & \multicolumn{1}{c}{\textcolor{red}{no}} & 
\multicolumn{1}{c}{\textcolor{red}{yes}} & 
\multicolumn{1}{c}{\textcolor{darkgreen}{white}} &
\multicolumn{1}{c}{\textcolor{darkgreen}{brick}}
\\\midrule
\rotatebox{90}{HAN+pair. (32)} & \multicolumn{1}{c}{\includegraphics[width=0.2\linewidth]{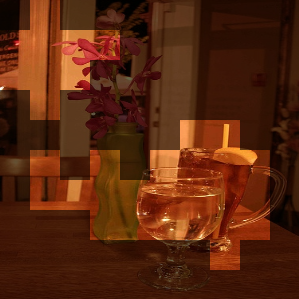}} &
\includegraphics[width=0.2\linewidth]{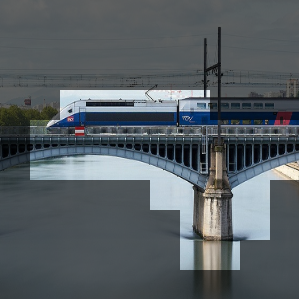} &
\includegraphics[width=0.2\linewidth]{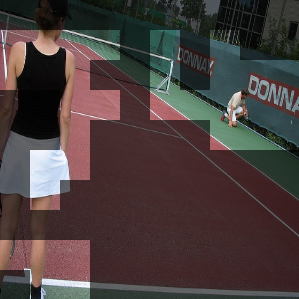} &
\includegraphics[width=0.2\linewidth]{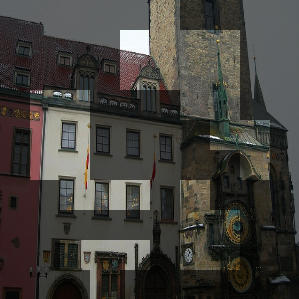}
\\
 & \multicolumn{1}{c}{\textcolor{darkgreen}{yes}} & 
\multicolumn{1}{c}{\textcolor{darkgreen}{no}} & 
\multicolumn{1}{c}{\textcolor{red}{black}} &
\multicolumn{1}{c}{\textcolor{orange}{stone}}
\\\midrule
\rotatebox{90}{HAN+pair. (16)} & \multicolumn{1}{c}{\includegraphics[width=0.2\linewidth]{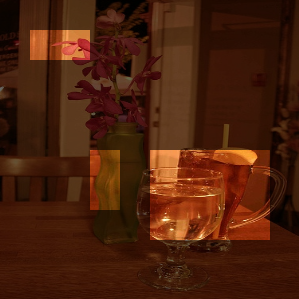}} &
\includegraphics[width=0.2\linewidth]{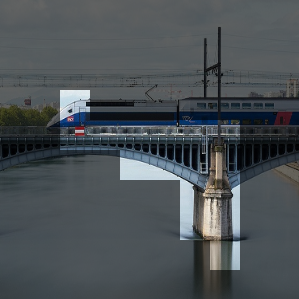} &
\includegraphics[width=0.2\linewidth]{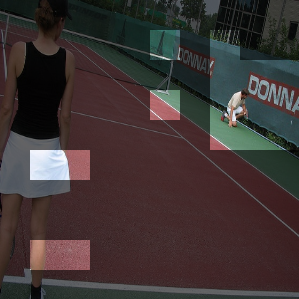} &
\includegraphics[width=0.2\linewidth]{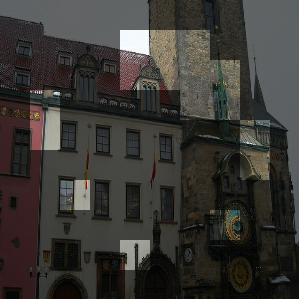}
\\
 & \multicolumn{1}{c}{\textcolor{darkgreen}{yes}} & 
\multicolumn{1}{c}{\textcolor{darkgreen}{no}} & 
\multicolumn{1}{c}{\textcolor{red}{blue}} &
\multicolumn{1}{c}{\textcolor{orange}{stone}}
\\\hline
\bottomrule

\end{tabular}
\end{center}
\caption{
Qualitative comparison between our variants of the hard attention mechanism together with different aggregation methods. The first row shows AdaHAN+pairwise (AdaHAN+pair. in the figure), the second row shows AdaHAN+sum, the third row shows HAN+pairwise with fixed $32$ entities, 
and the last row shows HAN+pairwise with fixed $16$  entities, covering 32\% and 16\% of the input respectively. In the images, attended regions are highlighted while unattended are darkened. The green denotes correct answers, the red incorrect, and orange denotes partial consensus between the human answers. This figure illustrates various strengths of the proposed methods. Best viewed on a display.
}
\label{fig:qualitative}
\end{figure*}

\begin{figure*}[p]
\begin{center}
\def\arraystretch{1}
\begin{tabular}{l@{\ }l@{\ }c@{\ }c@{\ }c@{\ }c}
 \toprule
 & \multicolumn{1}{c}{{\footnotesize Are this lions?}} & 
\multicolumn{1}{c}{{\footnotesize Is that}} & 
\multicolumn{1}{c}{{\footnotesize How many} } &
\multicolumn{1}{c}{{\footnotesize How many boats}} \\
& \multicolumn{1}{c}{} & 
\multicolumn{1}{c}{{\footnotesize an airplane}} & 
\multicolumn{1}{c}{{\footnotesize cars are on}} & 
\multicolumn{1}{c}{{\footnotesize are there?}} \\
& \multicolumn{1}{c}{} & 
\multicolumn{1}{c}{{\footnotesize  in image?}} & 
\multicolumn{1}{c}{{\footnotesize train tracks?}} & 
\multicolumn{1}{c}{}
\\
\rotatebox{90}{AdaHAN+pair.} & 
\includegraphics[width=0.2\linewidth]{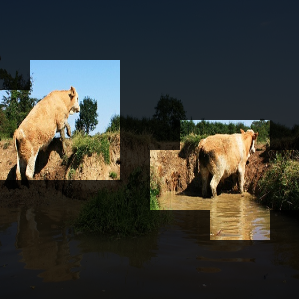} &
\includegraphics[width=0.2\linewidth]{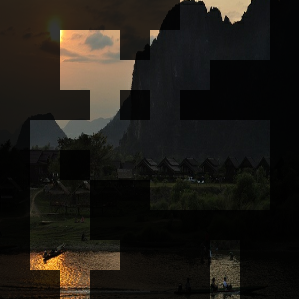} &
\includegraphics[width=0.2\linewidth]{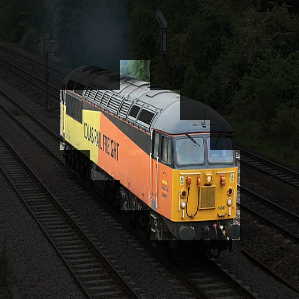} &
\multicolumn{1}{c}{\includegraphics[width=0.2\linewidth]{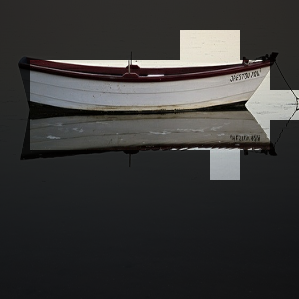}} 
\\
 & \multicolumn{1}{c}{\textcolor{darkgreen}{no}} & 
\multicolumn{1}{c}{\textcolor{darkgreen}{no}} & 
\multicolumn{1}{c}{\textcolor{red}{2} (1)} &
\multicolumn{1}{c}{\textcolor{red}{2} (1)}
  \\\midrule
 & \multicolumn{1}{c}{About how high} & 
 \multicolumn{1}{c}{Any chains on} &
\multicolumn{1}{c}{About how old} & 
\multicolumn{1}{c}{What color is} \\
& \multicolumn{1}{c}{is man jumping?} &
\multicolumn{1}{c}{hydrant?} &
\multicolumn{1}{c}{is boy on} & 
\multicolumn{1}{c}{this train?}  \\
& \multicolumn{1}{c}{} & 
\multicolumn{1}{c}{} & 
\multicolumn{1}{c}{surfboard?} & 
\multicolumn{1}{c}{}
\\
\rotatebox{90}{AdaHAN+pair.} & \multicolumn{1}{c}{\includegraphics[width=0.2\linewidth]{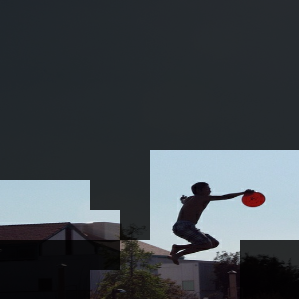}} &
\includegraphics[width=0.2\linewidth]{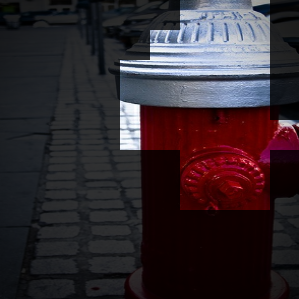} &
\includegraphics[width=0.2\linewidth]{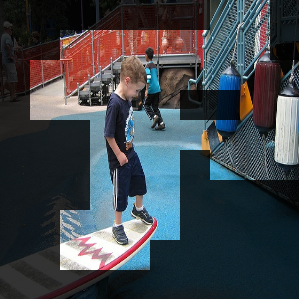} &
\includegraphics[width=0.2\linewidth]{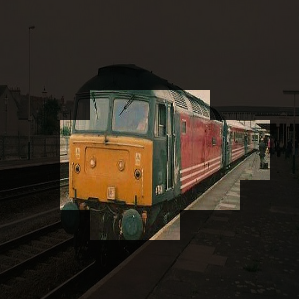}
\\
 & \multicolumn{1}{c}{\textcolor{darkgreen}{3 feet}} &
\multicolumn{1}{c}{\textcolor{darkgreen}{no}} &
\multicolumn{1}{c}{\textcolor{orange}{4}} & 
\multicolumn{1}{c}{\textcolor{orange}{yellow}} 
  \\\midrule
 & \multicolumn{1}{c}{Would this type} & 
 \multicolumn{1}{c}{What is} &
\multicolumn{1}{c}{What is he} & 
\multicolumn{1}{c}{What time does} \\
& \multicolumn{1}{c}{of train be used} &
\multicolumn{1}{c}{in his hand?} &
\multicolumn{1}{c}{holding up?} & 
\multicolumn{1}{c}{clock read?}  \\
& \multicolumn{1}{c}{as commuter?} & 
\multicolumn{1}{c}{} & 
\multicolumn{1}{c}{} & 
\multicolumn{1}{c}{}
\\
\rotatebox{90}{AdaHAN+pair.} & \multicolumn{1}{c}{\includegraphics[width=0.2\linewidth]{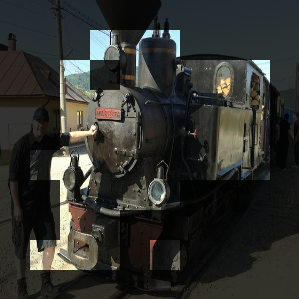}} &
\includegraphics[width=0.2\linewidth]{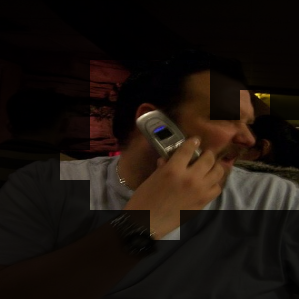} &
\includegraphics[width=0.2\linewidth]{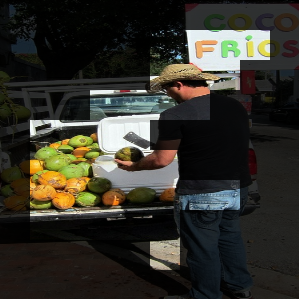} &
\includegraphics[width=0.2\linewidth]{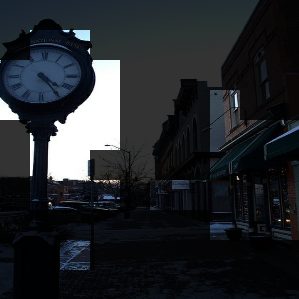}
\\
 & \multicolumn{1}{c}{\textcolor{darkgreen}{no}} &
\multicolumn{1}{c}{\textcolor{darkgreen}{phone}} &
\multicolumn{1}{c}{\textcolor{orange}{fruit}} & 
\multicolumn{1}{c}{\textcolor{red}{noon} (4:25)} 
  \\\midrule
 & \multicolumn{1}{c}{What is girl} & 
 \multicolumn{1}{c}{Are people } &
\multicolumn{1}{c}{Are all these} & 
\multicolumn{1}{c}{What kind} \\
& \multicolumn{1}{c}{trying to catch?} &
\multicolumn{1}{c}{sunbathing at} &
\multicolumn{1}{c}{people moving?} & 
\multicolumn{1}{c}{of animals}  \\
& \multicolumn{1}{c}{} & 
\multicolumn{1}{c}{beach in photo?} & 
\multicolumn{1}{c}{} & 
\multicolumn{1}{c}{are these?}
\\
\rotatebox{90}{AdaHAN+pair.} & \multicolumn{1}{c}{\includegraphics[width=0.2\linewidth]{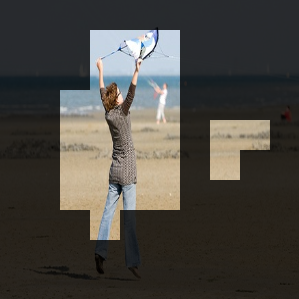}} &
\includegraphics[width=0.2\linewidth]{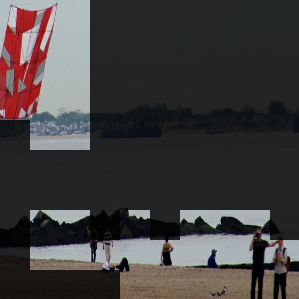} &
\includegraphics[width=0.2\linewidth]{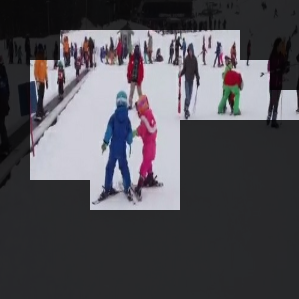} &
\includegraphics[width=0.2\linewidth]{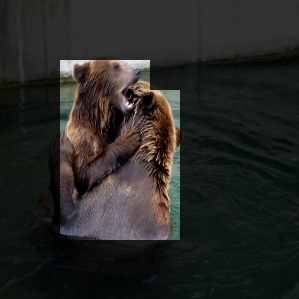}
\\
 & \multicolumn{1}{c}{\textcolor{darkgreen}{kite}} &
\multicolumn{1}{c}{\textcolor{darkgreen}{no}} &
\multicolumn{1}{c}{\textcolor{darkgreen}{yes}} & 
\multicolumn{1}{c}{\textcolor{orange}{bear}} 
\\\hline
\bottomrule

\end{tabular}
\end{center}
\caption{
We show additional results with our AdaHAN+pairwise. In the images, the attended regions are highlighted while the unattended are darkened. Green denotes correct, and red incorrect answers. Orange denotes partial consensus. Best viewed on display.
}
\label{fig:qualitative2}
\end{figure*}

One advantage of our formulation is that it is straightforward to visualize the masks of attended cells given questions and images, which we show in \autoref{fig:qualitative} and \autoref{fig:qualitative2}.
In general, relevant objects are usually attended, and that significant portions of the irrelevant background is suppressed. Although some background might be kept, we hypothesize the context matters in answering some questions.

In \autoref{fig:qualitative}, we show results with our different hard-attention mechanisms (HAN or AdaHAN), and different aggregation operations (summation or pairwise). We can see that the important objects are attended together with some context, which we hypothesize can also be important in correctly answering questions. These masks
are occasionally useful for diagnosing behavior. For instance, as row 2 and column 3 suggest, the network may answer the question correctly but likely for wrong reasons. 
We can also see broad differences between the network architectures.  For instance, the sum pooling method (row 2) is much more spatially constrained than the pairwise pooling version (row 1), even though the adaptive attention can select an arbitrarily large region.   We hypothesize that more visual features may unnecessarily interfere during the summation, and hence a more spatially sparse representation is preferred, or that sum pooling struggles to integrate across complex scenes.  The support is also not always contiguous: non-adaptive hard attention with 16 entities (row 4) in particular distributes its attention widely.

In \autoref{fig:qualitative2}, we show results with our best-performing model on VQA-CP: adaptive hard attention mechanism tied with a non-local, pairwise aggregation mechanism (AdaHAN+pairwise). The qualitative behaviour of this mechanism subsumes various fixed hard-attention variants, and with a variable spatial support tends to be better qualitatively and quantitatively than others. 
Interestingly, the topology of the attended parts of AdaHAN+pairwise differs from image to image. For instance, for the question ``Are this lions?'' (1st row, 1st column), the two attended regions are separated and quite localized.  However, for ``Is that an airplane in image?'' (1st row, 2nd column), the attended regions are contiguous and cover almost whole image.  The shape of the train in the image (1st row, 3rd column), despite of its elongated shape, is quite well captured by our method.  Similarly, we can observe that the attended regions overlap with the shape of a boat (1st row, 4th column), even though the method ultimately gets the question wrong.

\def\arraystretch{1.2}

\subsection{End-to-end Training.}
\label{sec:end2end_training}
Since our network uses hard attention, which has zero gradients almost everywhere, one might suspect that it will become more difficult to train the lower-level features, or worse, that untrained features might prevent us from bootstrapping the attention mechanism.
Therefore, we also trained HAN+sum (with 16\% of the input cells) end-to-end together with a relatively small convolutional neural network initialized from scratch.  
We compare our method against our implementation of the SAN method trained using the same simple convolutional neural network. 
We call the models: simple-SAN, and simple-HAN.
\newline
\noindent \textbf{Analysis.}
In our experiments, simple-SAN achieves about $21\%$ performance on the test set. Surprisingly, simple-HAN+sum achieves about $24\%$ performance on the same split, on-par with the performance of normal SAN that uses more complex and deeper visual architecture \cite{simonyan2014very}; the results are reported by \cite{agrawal2017don}. This result shows that the hard attention mechanism can indeed be tightly coupled within the training process, and that the whole procedure does not rely heavily on the properties of the ImageNet pre-trained networks.
In a sense, we see that a discrete notion of entities also ``emerges'' through the learning process, leading to efficient training.
\newline
\noindent \textbf{Implementation Details.}
In our experiments we use a simple CNN built of: 1 layer with 64 filters and 7-by-7 filter size followed up by 2 layers with 256 filters and 2 layers with 512 filters, all with 3-by-3 filter size. We use strides 2 for all the layers. 
\begin{figure}[!htb]
\begin{center}
\begin{subfigure}{.5\textwidth}
  \centering
  \includegraphics[width=0.8\linewidth]{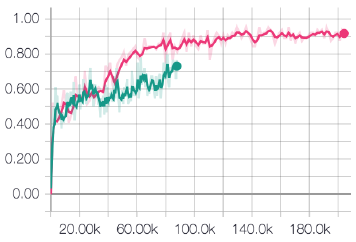}
  \caption{HAN+RN (purple), RN (green)}
  \label{fig:han_rn_12_hours}
\end{subfigure}%
\begin{subfigure}{.5\textwidth}
  \centering
  \includegraphics[width=0.87\linewidth]{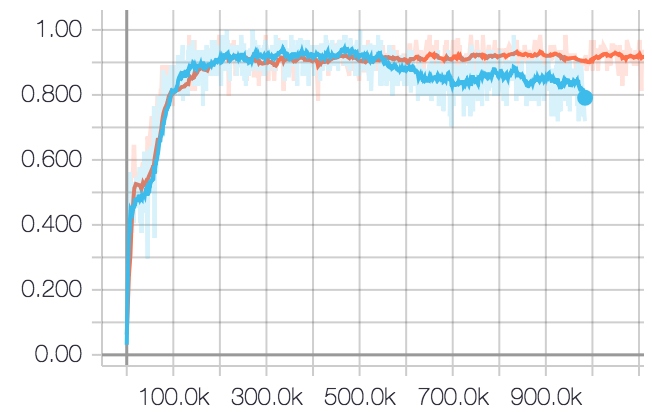}
  \caption{HAN+RN (orange), ST+RN (blue)}
  \label{fig:han_vs_st}
\end{subfigure}%
\end{center}
\caption{Validation accuracy plots on CLEVR of the methods under the same hyper-parameters setting~\cite{santoro2017simple}. (a)  HAN+RN ($0.25$ of the input cells)  and standard RN (all input cells) trained for 12 hours to measure the efficiency of the methods. (b) Our approaches to hard attention: the proposed one (orange), and the straight-through estimator (blue). }
\end{figure}

\subsection{CLEVR}
To demonstrate the generality of our hard attention method, particularly in domains that are visually different from the VQA images, 
we experiment with a synthetic Visual QA dataset termed CLEVR \cite{johnson2017clevr}, using a setup similar to the one used for VQA-CP and \cite{santoro2017simple}.  
Due to the visual simplicity of CLEVR, we follow up the work of~\cite{santoro2017simple}, and instead of relying on the ImageNet pre-trained features, we train our HAN+sum and HAN+RN (hard attention with relation network) architectures end-to-end together with a relatively small CNN (following~\cite{santoro2017simple}).
\newline
\noindent \textbf{Analysis.}
As reported in prior work~\cite{santoro2017simple,johnson2017clevr}, the soft attention mechanism used in SAN does not perform well on the CLEVR dataset, and achieves only $68.5\%$~\cite{johnson2017clevr} (or $76.6\%$~\cite{santoro2017simple}) performance. In contrast, relation network, which also realizes a non-local and pairwise computational model, essentially solves this task, achieving $95.5\%$ performance on the test set.
Surprisingly, our HAN+sum achieves 
89.7\%
performance even without a relation network, and HAN+RN (\ie, relation network is used as an aggregation mechanism) achieves 
$93.9\%$ on the test set.
These results show the mechanism can readily be used with other architectures on another dataset with different visuals. Training with HAN requires far less computation than the original relation network~\cite{santoro2017simple}, although performance is slightly below relation network's 95.5\%. \autoref{fig:han_rn_12_hours} compares computation time: HAN+RN and relation network are trained for 12 hours under the same hyper-parameter set-up. Here, HAN+RN achieves around 90\% validation accuracy, whereas relation network only 70\%.

Owing to hard-attention, we are able to train larger models, which we call HAN+sum$^{+}$, HAN+RN$^{+}$, and HAN+RN$^{++}$.
These models use larger CNN and LSTM, and HAN+RN$^{++}$ also uses higher resolution of the input (see Implementation Details below). The models
achieve 94.7\%, 96.9\% and 98.8\% respectively.
The relation network with hard attention operates on $k^2$ selected input cells, instead of all $n^2$ cells of the original RN. 
All our experiments except HAN+RN$^{++}$ use only one fourth of the input cells ($\frac{k}{n} = 0.25$, where $n = 64$). HAN+RN$^{++}$ uses a larger spatial tensor (14x14), and the same number of input cells as the original RN~\cite{santoro2017simple}, with $k = 64$ yielding ($\frac{k}{n} = \frac{64}{196})$ around 33\% of the input cells. Dealing with only the fraction of the input data helps the whole network to train faster.

\autoref{table:detailed_clevr} gives more context regarding the results on the CLEVR dataset that we are aware of, and compares our method with other approaches to answer questions about CLEVR images. Our best performing method, denoted by HAN+RN$^{++}$ that uses a deeper model and operates on larger input tensor than the original RN~\cite{santoro2017simple}, is very competitive to alternative approaches such us  FiLM~\cite{perez2017film}, TbD~\cite{mascharka2018transparency}, or MAC~\cite{hudson2018compositional}; and as \cite{santoro2017simple} and \cite{mascharka2018transparency} (TbD+hres) have noted increasing the spatial resolution definitely helps in achieving better performance.
As we can see in \autoref{table:detailed_clevr}, all the approaches seem to struggle with difficult counting questions, and RN is significantly worse on the {\it Compare Numbers} questions. In the remaining question types, HAN+RN$^{++}$ is either on par or even better than TbD+hres that uses larger spatial resolution, deep pre-trained image CNN, more specialized modules, and requires an ``expert layout''~\cite{explainable2018eccv}. 
Here, we keep the conceptual simplicity of the original RN~\cite{santoro2017simple} coupled with our simple mechanism of selecting important features, as well as we trained the whole architecture end-to-end and from scratch.
Finally, through a visual inspection, we have observed that the fraction of input cells that we have experimented with ($k=16$ for 8x8 spatial tensor, and $k=64$ for 14x14 spatial tensor) is sufficient to cover all the important objects in the image, and thus the mechanism resembles more the saliency mechanism. It is worth noting, the hard-attention mechanism often selects a few cells that correspond to the object as this is sufficient to recognize the object's properties such as size, material, color, type, and spatial location.

\noindent\textbf{Straight-Through Estimator.}
As an alternative to our hard attention, we have also implemented a few variants of the straight-through estimator~\cite{bengio2013estimating}, which is a method introduced to deal with non-differentiable neural modules.
In a nutshell, during the forward pass we employ steps that are non-differentiable or have gradients that are zero almost everywhere (e.g., hard thresholding), but in the backward pass, we introduce skip-connections that the back-propagation mechanism uses to bypass these steps.
For the purpose of gracefully implementing this mechanism in TensorFlow, we have implemented the estimator as follows\footnote{Credit goes to Sergey Ioffe for pointing out the general expression that we have adapted for our purpose.}. Let $\bs{x} \in \mathbb{R}^{n \times d}$ be spatial input, with $n$ spatial cells, each $d$-dimensional. All our estimators have the form
\begin{align*}
\bs{x} \cdot (g(\bs{x}) + \text{stop}(\mathbbm{1}\left\{g(\bs{x}) > t(g(\bs{x}), k)\right\} - g(\bs{x}) ))
\end{align*}
Here, $\cdot$ is the element-wise multiplication, $\text{stop}(\bs{y})$ prevents from propagating the gradient through $\bs{y}$,  $t(\bs{y},k)$ returns the $k$-th largest element of the vector $\bs{y}$, $\mathbbm{1}\left\{\mathcal{P}\right\}$ outputs 1 if the predicate $\mathcal{P}$ is true and 0 otherwise, and $g$ produces a spatial mask similar to the soft attention mask, i.e. $g(\bs{y}) \in \mathbb{R}^{n \times 1}$. In all our experiments, $g = \mu \circ f$ is the composition of the normalization function (e.g. softmax) $\mu$ and an MLP $f$ with one hidden layer of dimension $\frac{d}{2}$, and one ReLU between the hidden and the output layers. For $\mu$, we investigate identity, sigmoid or softmax. Only the latter two yield results significantly better than $60\%$, but we still find the results either under-performing to our hard-attention approach, or very unstable. For instance, \autoref{fig:han_vs_st} shows our best results (accuracy- and stability-wise) with straight-through. Moreover, our formulation of the straight-through still requires to have gradients back-propagated through all the cells, even though they are ignored in the forward-pass, and hence the method lacks the computational benefits of our hard-attention mechanism.

\noindent\textbf{Implementation Details.}
 In the experiments with HAN+Sum, and HAN+RN we follow the same setup as \cite{santoro2017simple}. However, we have made slight changes with our larger models: HAN+Sum$^{+}$, HAN+RN$^{+}$, and HAN+RN$^{++}$.
  HAN+Sum$^{+}$, HAN+RN$^{+}$, and HAN+RN$^{++}$ use an LSTM with 256 hidden units and 64 dimensional word embedding (jointly trained from scratch together with the whole architecture) for language.  For the image, we use a CNN with 4 layers, each with stride 2, 3x3 kernel size, ReLU non-linearities, and 128 features at each spatial cell.  Our classifier  is an MLP with a single hidden layer (1024 dimensional),  drop-out 50\%, and a single ReLU. Function $g_\theta$ defined in~\cite{santoro2017simple} is an MLP with four hidden layers (each 256 dimensional) and ReLUs. We also find that, before the sum-pooling in HAN+Sum$^{+}$, and before the pairwise aggregation in HAN+RN$^{+/++}$ it is worthwhile to process the multimodal embedding with a 1-by-1 convolution (we use 4 layers, with ReLUs, and 256 features). We use $l_2$-norm on all the weights as the regularization. For hard-attention, we have also found batch-normalization in the image CNN to be crucial to achieve a good performance. Moreover, batch-normalization before 1-by-1 convolutions is also helpful, but not critical. The other hyper-parameters are identical to the ones presented in~\cite{santoro2017simple}.

\begin{table}
\centering
\begin{threeparttable}
\begin{tabular}{l | C{14mm} | C{10mm} C{10mm} C{14mm} C{14mm} C{14mm}}
\toprule
Model & \textbf{Overall} & Count & Exist & Compare Numbers & Query Attribute & Compare Attribute \\
\hline
Human~\cite{johnson2017clevr}  & 92.6 & 86.7 & 96.6 & 86.5 & 95.0 & 96.0 \\
\hline
Q-type baseline~\cite{johnson2017clevr,antol2015vqa}  & 41.8 & 34.6 & 50.2 & 51.0 & 36.0 & 51.3  \\
LSTM-only~\cite{johnson2017clevr,ren2015image,antol2015vqa,gao2015you,malinowski2015ask} & 46.8 & 41.7 & 61.1 & 69.8 & 36.8 & 51.8  \\
CNN$+$LSTM~\cite{johnson2017clevr,ren2015image,antol2015vqa,gao2015you,malinowski2015ask}  & 52.3 & 43.7 & 65.2 & 67.1 & 49.3 & 53.0  \\
SAN~\cite{johnson2017clevr,yang2015stacked}       & 68.5 & 52.2 & 71.1 & 73.5 & 85.3 & 52.3  \\
SAN*~\cite{santoro2017simple,yang2015stacked}     & 76.6 & 64.4 & 82.7 & 77.4 & 82.6 & 75.4  \\
LBP-SIG~\cite{chen2017sva}  & 78.0 & 61.3 & 79.6 & 80.7 & 88.6 & 76.3  \\
N2NMN~\cite{hu2017learning}  & 83.7 & 68.5 & 85.7 & 85.0 & 90.0 & 88.9 \\
PG+EE (700k)$^-$~\cite{johnson2017inferring} & 96.9 & 92.7 & 97.1 & 98.7 & 98.1 & 98.9 \\
RN~\cite{santoro2017simple} & 95.5 & 90.1 & 97.8 & 93.6 & 97.9 & 97.1 \\
Hyperbolic RN~\cite{gulcehre2018hyperbolic} & 95.7 & - & - & - & - & - \\
Object RN**~\cite{desta2018object} & 94.5 & 93.6 & 94.7 & 93.3 & 95.2 & 94.4 \\
Stack-NMNs**~\cite{explainable2018eccv} & 96.6 (93.0) & - & - & - & - & - \\
FiLM~\cite{perez2017film} & 97.6 & 94.5 & 99.2 & 93.8 & 99.2 & 99.0 \\
DDRprog$^-$~\cite{suarez2018ddrprog} & 98.3 & 96.5 & 98.8 & 98.4 & 99.1 & 99.0 \\
MAC~\cite{hudson2018compositional} & 98.9 & 97.2 & 99.5 & 99.4 & 99.3 & 99.5 \\
TbD$^{-}$~\cite{mascharka2018transparency} & 98.7 & 96.8 & 98.9 & 99.1 & 99.4 & 99.2 \\
TbD+hres$^{-}$~\cite{mascharka2018transparency} & 99.1 & 97.6 & 99.2 & 99.4 & 99.5 & 99.6 \\
\hline
HAN+Sum$^{+}$ (Ours) & 94.7 & 88.9 & 97.3 & 88.0 & 98.1 & 97.0 \\
HAN+RN$^{+}$ (Ours) & 96.9 & 92.8 & 98.6 & 94.9 & 98.9 & 98.2 \\
HAN+RN$^{++}$ (Ours) & 98.8 & 97.2 & 99.6 & 96.9 & 99.6 & 99.6 \\
\bottomrule\addlinespace[1ex]
\end{tabular}
\end{threeparttable}
\caption{Results, in $\%$, on CLEVR. 
SAN denotes the SAN~\cite{yang2015stacked} implementation of \cite{johnson2017clevr}.
SAN* denotes the SAN implementation of \cite{santoro2017simple}.
Object RN**~\cite{desta2018object} and Stack-NMNs**~\cite{explainable2018eccv} report the results only on the validation set, whereas others report on the test set. Overall performance of Stack-NMNs**~\cite{explainable2018eccv} is measured with the ``expert layout'' (similar to N2NMN) yielding 96.6 and without it (93.0).
DDRprog$^-$~\cite{suarez2018ddrprog}, PG+EE (700k)$^-$~\cite{johnson2017inferring}, TbD$^-$, and TbD+hres$^-$~\cite{mascharka2018transparency} are trained with a privileged state-description, while others are trained directly from images-questions-answers. TbD+hres~\cite{mascharka2018transparency} uses high-resolution (28x28) spatial tensor, while majority uses either 8x8 or 14x14. 
HAN+Sum/RN$^{+}$ denotes a larger relational model, or a different hyper-parameters setup, than the model of \cite{santoro2017simple}. 
HAN+RN$^{++}$ denotes HAN+RN$^{+}$ with larger input images with spatial dimensions 224x224 as opposed to 128x128, and larger image tensors with spatial dimension 14x14 as opposed to 8x8. 
}
\label{table:detailed_clevr}
\end{table}
\section{Summary}
\label{sec:summary}
We have introduced a new approach for hard attention in computer vision that 
selects a subset of the 
feature vectors for further processing based on the their magnitudes.
We explored two models, one which selects subsets with a pre-specified number of vectors (HAN), and the other one that adaptively chooses the subset size as a function of the inputs (AdaHAN). 
Hard attention is often avoided in the literature because 
it poses a challenge for gradient-based methods due to non-differentiability. 
However, since we found our feature vectors' magnitudes correlate with relevant information, our hard attention mechanism exploits this property to perform the selection. 
Our results showed our HAN and AdaHAN gave competitive performance on challenging Visual QA datasets. 
Our approaches 
seem to be at least as good as
a more commonly used soft attention mechanism while providing additional computational efficiency benefits.
This is especially important for the increasingly popular class of non-local approaches, which often require computations and memory which are quadratic in the number of the input vectors. Finally, our approach also provides interpretable representations, as the spatial locations of the selected features correspond most strongly to those parts of the image which contributed most strongly.

\subsection*{Acknowledgments}
We would like to thank Aishwarya Agrawal, Relja Arandjelovic, David G.T. Barrett, Joao Carreira, Timothy Lillicrap, Razvan Pascanu, David Raposo, and many others on the DeepMind team for critical feedback and discussions.

\clearpage

\bibliographystyle{splncs}
\bibliography{biblioLong,egbib}
\end{document}